\documentclass[letterpaper, 10 pt, conference]{ieeeconf}  

\IEEEoverridecommandlockouts                              

\usepackage{graphicx} 
\usepackage{epstopdf}
\DeclareGraphicsExtensions{.eps}

\usepackage{multirow}
\usepackage{algorithmic}
\usepackage{algorithm}
\usepackage{amsmath} 
\usepackage{amssymb}
\usepackage{amsxtra}
\usepackage{url} 
\usepackage{color} 
\usepackage{bm}
\usepackage{placeins}
\usepackage[caption=false]{subfig}
\usepackage{xspace}
\usepackage{rotating}
\usepackage{siunitx}
\usepackage{textcomp}%
\usepackage{algorithm}
\usepackage{algorithmic}

\newcommand{\myvector}[1]{\bm{#1}}
\newcommand{\myvec}[1]{\myvector{#1}}

\newcommand{\R}[1]{\mathbb{R}^{#1}}

\newcommand{\argmax}{\operatornamewithlimits{arg max}}

\newcommand{\excise}[1]{}

\newif\ifremark
\long\def\remark#1{
  \ifremark%
  \begingroup%
  \dimen0=\textwidth
  \advance\dimen0 by -1in%
  \setbox0=\hbox{\parbox[b]{\dimen0}{\protect\em #1}}
  \dimen1=\ht0\advance\dimen1 by 2pt%
  \dimen2=\dp0\advance\dimen2 by 2pt%
  \vskip 0.25pt%
  \hbox to \textwidth{%
    \vrule height\dimen1 width 3pt depth\dimen2%
    \hss\copy0\hss%
    \vrule height\dimen1 width 3pt depth\dimen2%
  }%
  \endgroup%
  \fi}

\newcommand{\X}{O}
\newcommand{\x}{\ensuremath{\myvec{o}}}

\newcommand{\uv}{\ensuremath{a}}
\newcommand{\ui}[1]{a}

\newcommand{\ac}{\uv}

\usepackage{graphicx} 
\usepackage{amsmath} 

\newcommand{\net}{FollowNet}
\newcommand{\trainsuccess}{54\%}
\newcommand{\devsuccess}{52\%}
\newcommand{\trainsuccesspartial}{70\%}
\newcommand{\devsuccesspartial}{61\%}
\newcommand{\trainnosuccess}{30\%}
\newcommand{\devnosuccess}{39\%}

\newcommand{\devsuccessnoatt}{40\%}

\newcommand{\devnosuccessnoatt}{45\%}

\title{\LARGE \bf
\net: Robot Navigation by Following Natural Language Directions with Deep Reinforcement Learning
}

\author{Pararth Shah$^{1}$ \and Marek Fiser$^{1}$ \and Aleksandra Faust$^{1}$ \and J. Chase Kew$^{1}$ \and Dilek Hakkani-Tur$^{1}$
\thanks{$^{1}$The authors are with Google, Mountain View, CA, USA
        {\tt\small \{pararth,mfiser,faust,jkew,dilekh\}@google.com}}%
}

\begin{document}

\maketitle
\thispagestyle{empty}
\pagestyle{empty}

\begin{abstract}
Understanding and following directions provided by humans can enable robots to navigate effectively in unknown situations. We present \net,~ an end-to-end differentiable neural architecture for learning multi-modal navigation policies. \net~ maps natural language instructions as well as visual and depth inputs to locomotion primitives. \net~ processes instructions using an attention mechanism conditioned on its visual and depth input to focus on the relevant parts of the command while performing the navigation task. Deep reinforcement learning (RL) a sparse reward learns simultaneously the state representation, the attention function, and control policies. We evaluate our agent on a dataset of complex natural language directions that guide the agent through a rich and realistic dataset of simulated homes. We show that the \net~ agent learns to execute previously unseen instructions described with a similar vocabulary, and successfully navigates along paths not encountered during training. The agent shows 30\% improvement over a baseline model without the attention mechanism, with \devsuccess~ success rate at novel instructions. \end{abstract}

\section{Introduction}

Humans often navigate unknown environments by observing their surroundings and following directions. These directions consist predominantly of landmarks and directional instructions and other common words. For example, humans can find a kitchen in a home they haven't visited before, by following directions such as: ``Turn right at the dining table, then take the second left''. This process requires visual observations, e.g. a dining table in the field of view or knowledge of a typical hallway, and execute actions present in the direction: turn left. There are multiple dimensions of complexity: limited field of view, qualifier words like ``second'', synonyms such as ``taking'' and ``turning'', understanding that ``take the second left'' refers to the door, etc.

\begin{figure}[tb]
	\begin{center}
		\begin{tabular}{cc}
			\subfloat[House 1]{\includegraphics[height=3.8cm,keepaspectratio=true,trim={0.5in 2.0in 0.5in 0},clip]{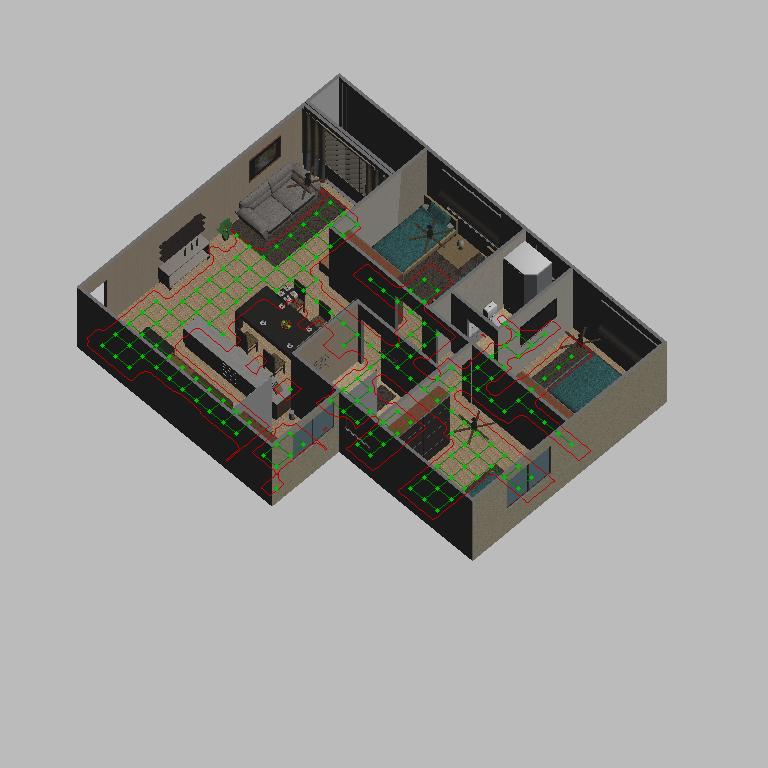}\label{fig:h1}} &
			\subfloat[House 2]{\includegraphics[height=3.8cm,keepaspectratio=true,trim={0.5in 1.5in 0.5in 0.5in},clip]{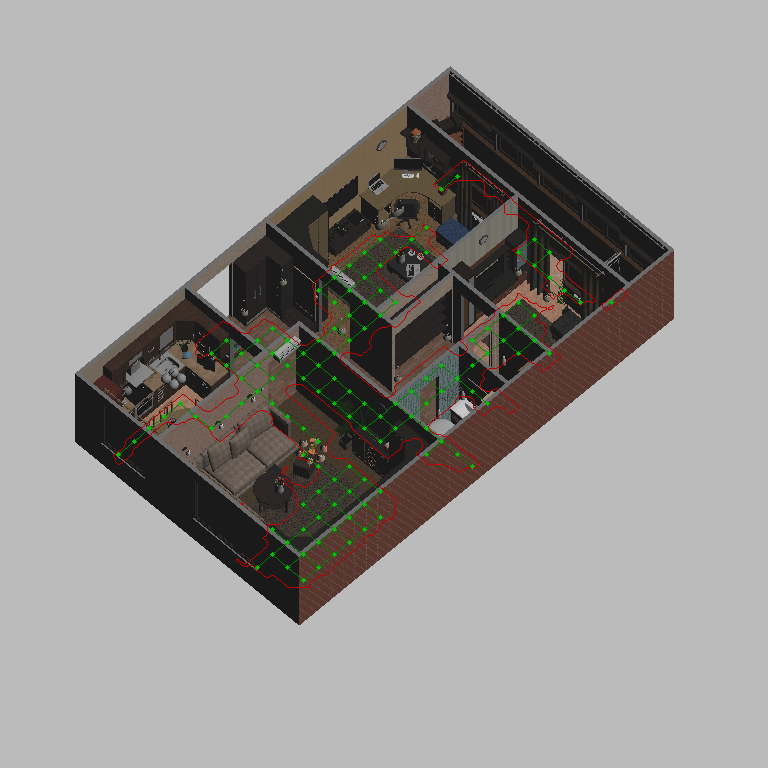}\label{fig:h3}}\\
		\end{tabular}
		\caption{3-dimensional rendering of the houses used for learning navigation from natural language instructions.}
		\label{fig:env}
	\end{center}
\end{figure}

\begin{figure*}[ht!]
  \begin{center}
  \includegraphics[width=0.95\linewidth]{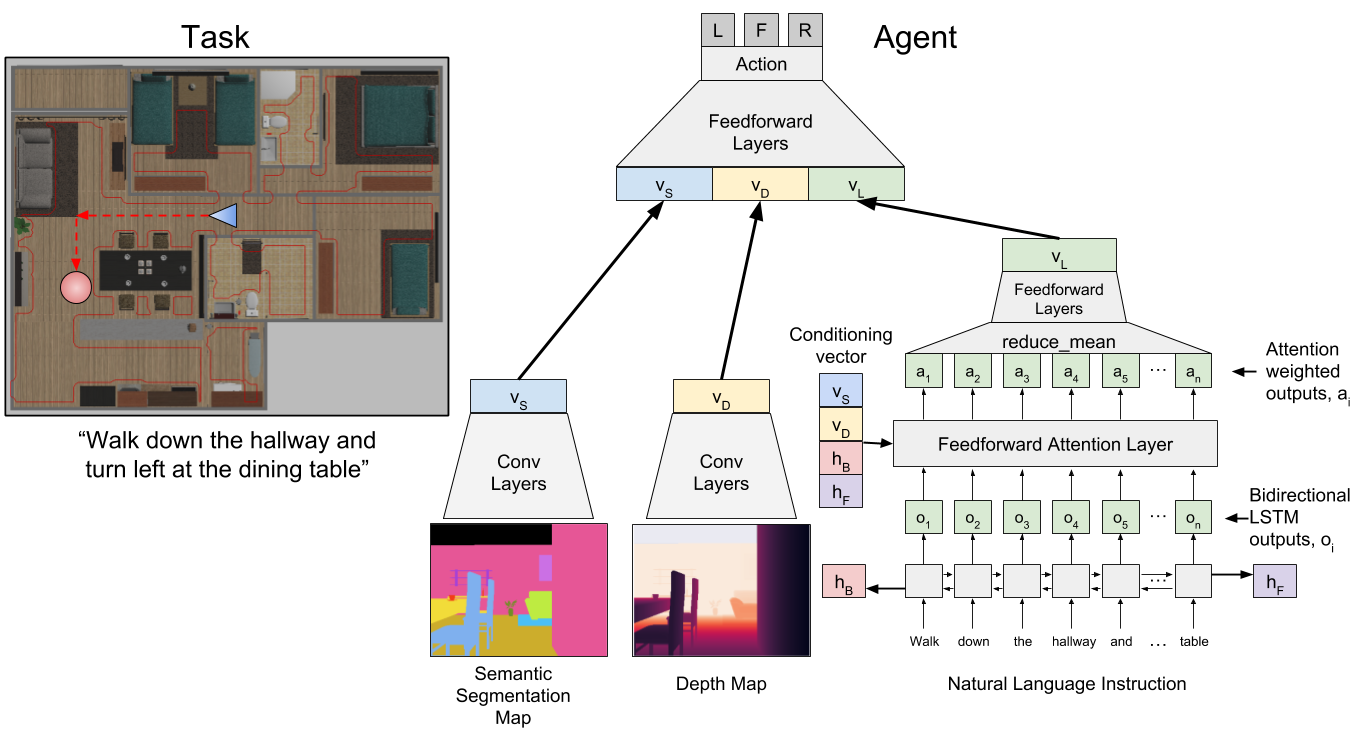}
  \caption{Neural model for mapping visual and language inputs to a navigation action. Left: An example task, where the robot starts at the position and orientation specified by the blue triangle, and must reach the goal location specified by the red circle. The robot receives a natural language instruction to follow the path marked in red, listed below the image. Right: the \net~ architecture. Semantic segmentation map is fed into a 3-layer convolutional net with 3, 8, and 16 outputs, [1,1], [4,4], [3,3] kernels, and 1, 2, 1 strides. The depth image is an input to 2-layer convolutional network with 8 and 16 outputs, [4, 4] and [3, 3] kernels, and 2 and 1 strides. The command is an input to a bidirectional GRU with 32 outputs. The Feedforward Attention Layer has soft attention 16 hidden states. Lastly the the Feedforward layers consist of two layers with 16 and 8 hidden units.}
  \label{fig:model}
  \end{center}
\end{figure*}

In this paper, we apply human-like direction following to robots navigating in 2-dimensional workspaces (Fig. \ref{fig:env}). We present robots with example directions similar to the one above, and train a deep reinforcement learning (DRL) agent to follow the directions. The agent is tested on how well it follows new directions when starting from different locations. We accomplish this with a novel deep neural net architecture, \net~(Fig. \ref{fig:model}), which is trained with Deep Q-Network (DQN) \cite{atari-paper}. The observation space consists of natural language instructions and visual and depth observations from the robot's vantage point (Fig. \ref{fig:input_table}). The policy's output is the next motion primitive to perform.  The robot moves along an obstacle-free grid, but the instructions require the robot to move over a variable number of nodes to reach the destination. The instructions we use (Table \ref{tab:training}) contain implicitly encoded rooms, landmarks, and motion primitives. In the example above, ``kitchen'' is the room that serves as the goal location. ``Dining table'' is an example of a landmark, a point at which the agent might change direction. Both rooms and landmarks are mapped to groups of grid points without the agent's knowledge. We use a sparse reward, given to the agent only when it reaches a waypoint. 

The novel aspect of the \net~ architecture is a language instruction attention mechanism that is conditioned on the agent's sensory observations. This allows the agent to do two things. First, it keeps track of the instruction command and focuses on different parts as it explores the environment. Second, it associates motion primitives, sensory observations, and sections of the instruction with the reward received, which enables the agent to generalize to new instructions.

We evaluate how well the agent generalizes to new instructions and new motion plans. First, we evaluate the agent on how well it follows previously unseen two-step directions in houses with which it is familiar. The results show that the agent follows \devsuccess~ directions completely and  \devsuccesspartial~ partially, a 30\% increase over a baseline. Second, the same instructions are valid for a set of different starting positions. For example, "Exit the room" is valid for any start location inside the room, yet the motion plan that the robot needs to execute to complete the task can be very different. To access how well the motion plans generalize to new start locations, we evaluate the agent on the instructions on which it was trained (up to five-step directions), but from new starting positions. The agent completes \trainsuccesspartial~ directions partially and \trainsuccess~ fully. To put that in perspective, multi step directions are challenging for people to perform as well.

\section{Related Work}

End-to-end navigation methods \cite{vin,prm-rl,vis-nav-target} use deep reinforcement learning on robots' sensory observations and relative goal location. In this work, we provide natural language instructions instead of the explicit goal, and the agent must learn to interpret the instructions to complete the task.
One challenge in reinforcement learning applied to robotics is the state space representation. Large state spaces slow down the learning, so often different approximation techniques are used. Examples of these are probabilistic roadmaps (PRMs) \cite{prm-rl,malone-journal-14} and simple discretization of the space \cite{nn-pursuit,nowe-rlartMulti-12}. Here, we discretize the 2-dimensional workspace and allow the agent to move through the grid from node to node. Essentially, we assume that the robot can avoid obstacles and move safely between two grid points by executing the motion primitive corresponding to the action.  

Deep learning has shown great success with learning natural language \cite{mesnil2015using, mikolov2013distributed} and vision \cite{alexnet,resnet} and even combining visual and language learning \cite{thomason-ijcai17,ispy}. 
Applied to robot motion planning and navigation, language learning typically requires some level of parsing with formal descriptions \cite{Matuszek2013}, semantic parsing \cite{robot-dialog-stone}, a probabilistic graphical model \cite{tellex11}, encoding and alignment \cite{mei2016navigational}, or task grounded language \cite{gated-language} etc. Learning object labels through natural language, though, has been addressed mainly by learning to parse natural language instructions into a hierarchical structure which can be used during planning and execution of robot actions~\cite{tellex2011understanding, thomason2015learning,granularities17}
and active learning \cite{thomason-corl17}. Here, similarly to \cite{Vision-and-Language}, we aim to implicitly learn the labels for landmarks (objects) and motion primitives (actions) and their interpretation with respect to visual observations. Unlike \cite{Vision-and-Language}, we use DQN \cite{atari-paper} over the proposed \net~to learn the navigation policy. Other works \cite{grounded-language} have used curriculum to complete several tasks in an environment. 

Another recent work that combines 3D navigation, vision, and natural language is learning to answer questions \cite{qa}.  The questions come from a prescribed set of questions where certain keywords are replaced. In our work, the language instructions given to the agent here are independently created by four people, and presented to the agent without any processing. 
Several methods learn from unfiltered language \cite{misra,interactive-language} and visual input. In these methods that visual input is an image of an entire planning environment. In contrast, FollowNet only receives partial environment observation.  

\section{Methods}
\subsection{Problem formulation}
\begin{figure}[]
	\begin{center}
		\begin{tabular}{ccc}
			\subfloat[Couch in living room]{\includegraphics[width=0.13\textwidth,keepaspectratio=true]{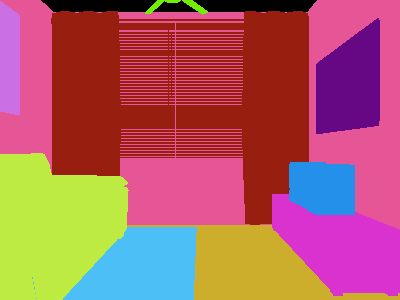}\label{fig:input_cauch}} &
			\subfloat[Table in front]{\includegraphics[width=0.13\textwidth,keepaspectratio=true]{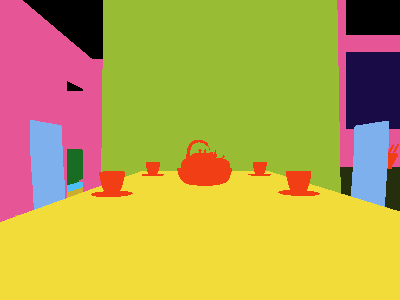}\label{fig:input_table}} &
			\subfloat[Table to left, couch to the right]{\includegraphics[width=0.13\textwidth,keepaspectratio=true]{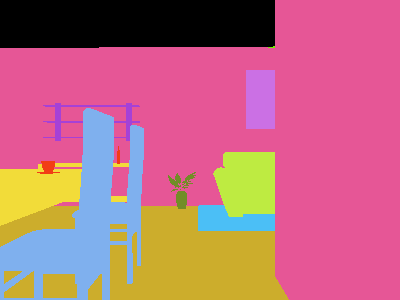}\label{fig:input_table_left}}\\
		\end{tabular}
		\caption{Semantic segmentation map observations for \net~ agents. Colors correspond to object types (unknown to the agent), and are consistent between the houses and vantage points. Couch is green (a and c), dinning table is yellow (b and c).}
		\label{fig:inouts}
	\end{center}
\end{figure}

We assume the robot to be a point-mass with three degrees of freedom ($x$, $y$, $\theta$), navigating in a 2-dimensional grid overlaid on a 3-dimensional indoor house environment (Fig. \ref{fig:env}). To train a DQN \cite{atari-paper} agent, we formulate the task as a Partially Observable Markov Decision Process (POMDP): a tuple $(O, A, D, R)$ with observations $\x = [\x_{NL}\; \x_V] \in O$, where $\x_{NL} = [w_1 w_2 \dots w_n]$ is a natural language instruction sampled from a set of user-provided directions for reaching a goal. The location of the goal is unknown to the agent. $\x_V$ is the visual input available to the agent, which consists of the image that the robot sees (Fig. \ref{fig:input_table_left}) at a time-step $i$. The set of actions $A = \{turn \frac{\pi}{2},\, go\ straight,\, turn \frac{3\pi}{2}\}$ enables the robot to either turn in place or move forward by a step. The system dynamics $D: \X \times A \rightarrow \X$ are deterministic and apply the action to the robot. The robot either transitions to the next grid cell or changes its orientation. Note, that the agent does not know where it is located in the environment. 

The reward $R:\X \rightarrow \R{}$ rewards an agent reaching a landmark (waypoint) mentioned in the instruction, with a reward of $+1.0$ if the waypoint is the final goal location, and a smaller reward of $+0.05$ for intermediate waypoints. The agent is rewarded only once for each waypoint in the instruction it reaches, and the episode terminates when the agent reaches the final waypoint, or after a maximum number of steps. Our aim is to learn an action-value function $Q: \X \rightarrow \R{\|A\|},$ approximated with a deep neural network and trained with DQN. 

Fig. \ref{fig:model} provides an example task, where the robot starts at the position and orientation specified by the blue triangle, and must reach the goal location specified by the red circle. The robot receives a natural language instruction (Table \ref{tab:training}) to follow the path marked in red.

\begin{table*}
\caption {Examples of instructions used in training. House \# is identity of the house, start and goal determine the valid regions of the instructions. The robot only has access to the instruction and visual observation, without the context of where it is located (house \#, start and goal) and the valid regions for the instruction. House \#, start and goal are used for the RL reward design.}
\begin{center}
\footnotesize
\begin{tabular}{l p{1.7cm} p{1.2cm} p{12cm}}
\label{tab:training}

House \# & Start & Goal & Instruction \\ \hline
1	&	Kids Bedroom	&	Bedroom	&	Exit the room, turn left and walk through the hall and enter the doorway in front of you	\\
1	&	Kitchen	&	Couch	&	Go out the door, to the opposite corner of the hallway, and go through the door. Then go to the opposite corner of the room.	\\
1	&	Bathroom	&	TV	&	Go out the door and forward until you see the plant. Then turn left and go through the door in front of you. Turn left and you should see the tv.	\\
2	&	Study	&	Gym	&	Go out the door and turn left. Go forward until you reach a doorway, then turn left. Go forward and through the door in front of you. Go straight through the bedroom and through the door on the far side.	\\
2	&	Gym	&	Kitchen	&	Go out the door, straight through the bedroom and out the door on the far side. Continue straight until you hit the wall, then turn right. After the corner of the wall, turn left and go through the door ahead.	\\
2	&	Hallway	&	Gym	&	Go out through the bedroom, straight across and through the far door.	\\
2	&	Couch in Living Room	&	Bedroom	&	Go out the door and turn right, then right again, then left at the bathroom. Go straight ahead and through the doorway in front of you.	\\
2	&	Dining Table	&	Bathroom	&	Go out the door and turn right after the corner of the wall. Go straight ahead and through the door.	\\
2	&	Bathroom	&	TV	&	Go out the door and turn right, then forward and through the door ahead of you. Go forward and turn left to reach the tv.	\\
2	&	Kitchen	&	Gym	&	Go out the door and straight across the hallway until you reach a doorway, then turn right. Go straight until you reach another doorway, then turn left. Go straight forward until you reach another doorway and go through that one. Go straight across the bedroom and through the door on the far side.	\\
\end{tabular}
\end{center}
\end{table*}

\subsection{\net}
We present \net, a neural architecture for approximating the action value function directly from the language and visual inputs (Fig. \ref{fig:model}). To simplify the image processing task, we assume a separate preprocessing step parses the visual input $\x_V \in \R{n \times m}$ to obtain a semantic segmentation $\x_S$ which assigns a one-hot semantic class id to each pixel, and a depth map $\x_D$ which assigns a real number to each pixel corresponding to the distance from the robot. The agent takes the ground truth $\x_S$ and $\x_D$ from its current point of view and runs each through a stack of convolutional layers followed by a fully-connected layer. From these it obtains fixed length embedding vectors $v_S \in \R{d_S}$ and $v_D \in \R{d_D}$ (where $d_X = length(v_X)$) that encode the visual information available to the agent.

We use a single layer bi-directional GRU network \cite{cho2014learning} with state size $d_L$ and initial state set to 0, to encode the natural language instruction using the following equations:
\begin{align*}
h_F, \{o_{i, F}\} & = GRU_F(\{w_i\}) \\
h_B, \{o_{i, B}\} & = GRU_B(\{w_i\}) \\
o_i & = [o_{i, F}\, o_{i, B}]
\end{align*}
where $h_F, h_B \in \R{d_L}$ are the final hidden states of the forward and backward GRU cells, respectively, while $o_i \in \R{2d_L}$ are the concatenated outputs of the forward and backward cells, corresponding to the embedded representation of each token conditioned on the entire utterance. To enable the agent to focus on different parts of the instruction depending on the context, we add a feed-forward attention layer over $o_i$:
\begin{align*}
v_C & = [v_S \; v_D \; h_B \; h_F] \\
e_i & = FF_A(v_C, o_i) \\
\alpha_i & = softmax(e_i) \\
a_i & = \alpha_i \; o_i \\
v_A & = (1/k) \Sigma_i^k a_i \\
v_L & = FF_L(v_A) \\
Q(\x) & = FF_Q([v_S, v_D, v_L])
\end{align*}
We use a feed-forward attention layer $FF_A$ conditioned on $v_C$, which is the concatenated embeddings of the visual and language inputs, to obtain unnormalized scores $e_i$ for each token $w_i$. $e_i$ are normalized using the softmax function to obtain the attention scores $\alpha_i$, which correspond to the relative importance of each token of the instruction for the current time step. We take the attention-weighted mean of the output vectors $o_i$, and pass it through another feed-forward layer to obtain $v_L \in \R{d_L}$, which is the final encoding of the natural language instruction.

The Q function is then estimated from the concatenated $[v_S, v_D, v_L]$ passed through a final feed-forward layer. During training, we sample actions from the Q-function using an epsilon-greedy policy to collect experience, and update the Q-network to minimize the Bellman error over batches of transitions using gradient descent. After the Q function is trained, we used the greedy policy $\pi(\x): \X \rightarrow A,$ with respect to learned $\hat Q,$ $\pi(\x) = \pi^{\hat Q}(\x) = \argmax_{\ac \in A} \hat Q(\x, \ac),$ to take the robot to the goal presented in the instruction $\x_l.$

\begin{figure}[]
	\begin{center}
		\begin{tabular}{cc}
			\subfloat[House 1]{\includegraphics[height=3.9cm,keepaspectratio=true]{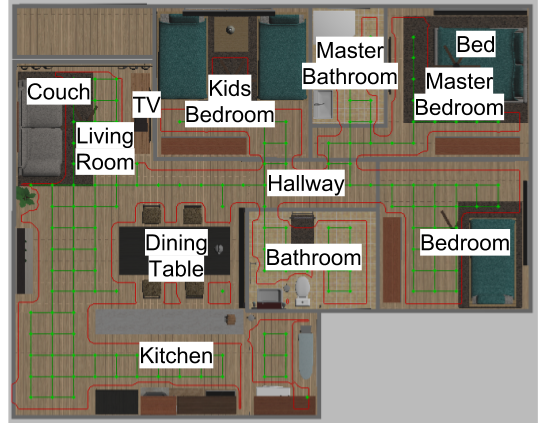}\label{fig:h1_labeled}} &
			\subfloat[House 2]{\includegraphics[height=3.9cm,keepaspectratio=true]{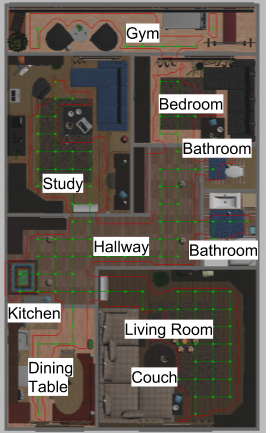}\label{fig:h3_labeled}}\\
		\end{tabular}
		\caption{Landmarks and grid overlaid over the environments.}
		\label{fig:env_labeled}
	\end{center}
\end{figure}

\section{Results}
In this Section we present the training and evaluation setup, and then evaluate \net~ against a baseline model without attentional layer. We also look into the effect of the attentional layer and task complexity. 
\subsubsection*{Setup and methodology}
We chose two houses from the SUNCG \cite{suncg} dataset that had many rooms and objects in common (Fig. \ref{fig:env}). The size of the grid for House 1 is $23 \times 18$ nodes, for House 2 $14 \times 20$ nodes.

For both houses we chose 7 navigation tasks consisting of starting and ending locations (e.g. ``Study to table in the kitchen''). Three people each independently wrote one instruction for each task in each house forward, and one instruction for the same task reversed (e.g. ``exit the room and take the door on the opposite wall''). After discarding some instructions for containing vocabulary not seen elsewhere, we settled on a set of 58 instructions. Examples of the tasks and instructions in the Table \ref{tab:training}. Each instruction contains implicitly stated waypoints. In the example above, the set of waypoints might be: study, table, kitchen, and door. The agent has no knowledge of the waypoints, and they are only used for reward computation and evaluation. 
We overlaid a navigation grid (1 meter edges) onto each house (Fig. \ref{fig:env_labeled}). For each instruction, we associated grid nodes with valid starting points and waypoints that we expect agent to reach when completing the instruction.  For each task, we selected approximately 5-10 starting nodes, where were randomly selected during the training. 

For the evaluation, we followed the same methodology to create an additional dataset of 15 instructions which introduce new combinations of instructions and locations not present in training. For example: ``Go out the door and straight across the hallway, then through the door in front of you.'' We made sure the evaluation instruction uses the same landmark and directional vocabulary as the training set without introducing new words. The evaluation instructions consists of two-step instructions, while the training set contains up to five-step instructions. The evaluation dataset consists of 100 queries, created over the 15 evaluation instructions and randomly sampled start and goal location within the start and goal area applicable to the instruction. For example, for Kids Bedroom anywhere in the room is the possible location to start an episode.

We trained \net~ agent with the learning rate $\alpha = 1.70974 * 10^{-4}$ and discount factor $\gamma = 0.990022,$ selected through a hyper-parameter tuning \cite{vizier}. We stop the training after $2\,500\,000$ steps. 
We compare \net~ to a baseline, which is an identical network but without the attention layers. The baseline consists of convolutional layers, RNN, and FCC layers. The baseline was trained and tuned in the same manner as the \net. It is a non-trivial and challenging baseline.

\begin{figure}[]
	\begin{center}
		\begin{tabular}{c}
        \subfloat[Performance on training set.]{\includegraphics[width=0.4\textwidth,height=3.8cm,keepaspectratio=false]{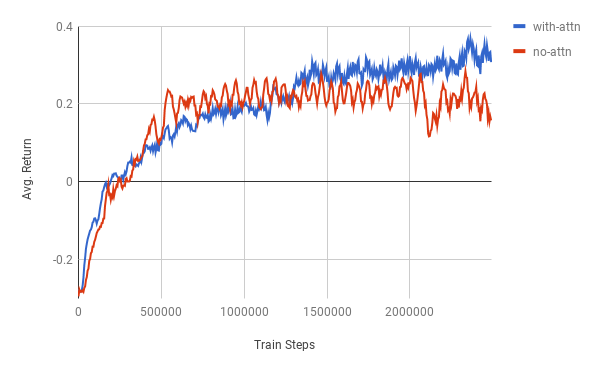}\label{fig:TrainReward}} \\
			\subfloat[Performance on hold-out set.]{\includegraphics[width=0.35\textwidth,height=3.8cm,keepaspectratio=false]{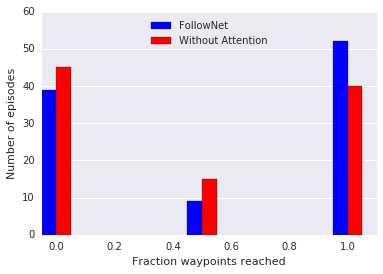}\label{fig:WaypointCompare}} 
		\end{tabular}
        \caption{Comparing \net~ (blue) with a baseline agent with no attention (red). Top: Average return on the training set plotted against no. of training steps. Evaluations done every 10,000 steps on the same hold out set. Bottom: Evaluated trained policies of both agents on a hold-out set for 100 episodes, and plotted a histogram of fraction of waypoints reached successfully. Learning to attend over the input instruction shows a 30\% relative increase (\devsuccess~ vs. \devsuccessnoatt) in fully successful episodes.}
		\label{fig:reward}
	\end{center}
\end{figure}

\subsubsection*{Comparison to baseline with no attention}
Fig. \ref{fig:reward} shows the learning curve for the \net~ and baseline over the holdout dataset. Early  in the training the model with no attention is showing a slightly better learning curve. With prolonged training, \net~ outperforms the baseline agent. This is because encoding the instruction with an RNN enables the agent to consider the relative ordering of words in the instruction. The visual inputs and language input are embedded separately and then fused into a single context vector which conditions the final action selection policy. Without attention on the RNN, the agent cannot selectively focus on parts of the instruction relevant to the visual context.

Fig. \ref{fig:WaypointCompare} depicts the histogram of fraction of waypoints reached successfully on the evaluation dataset. The \net~ agent's overall success at following instructions is \devsuccess~ on the evaluation dataset, while the baseline completes only \devsuccessnoatt. This means that the \net~ has 30\% relative increase over the baseline. We also see that \net~ has fewer fully unsuccessful tasks (\devnosuccess~ vs \devnosuccessnoatt~), that is 13\% relative decrease. 

\begin{figure}[]
	\begin{center}
		\includegraphics[width=0.35\textwidth,keepaspectratio=true]{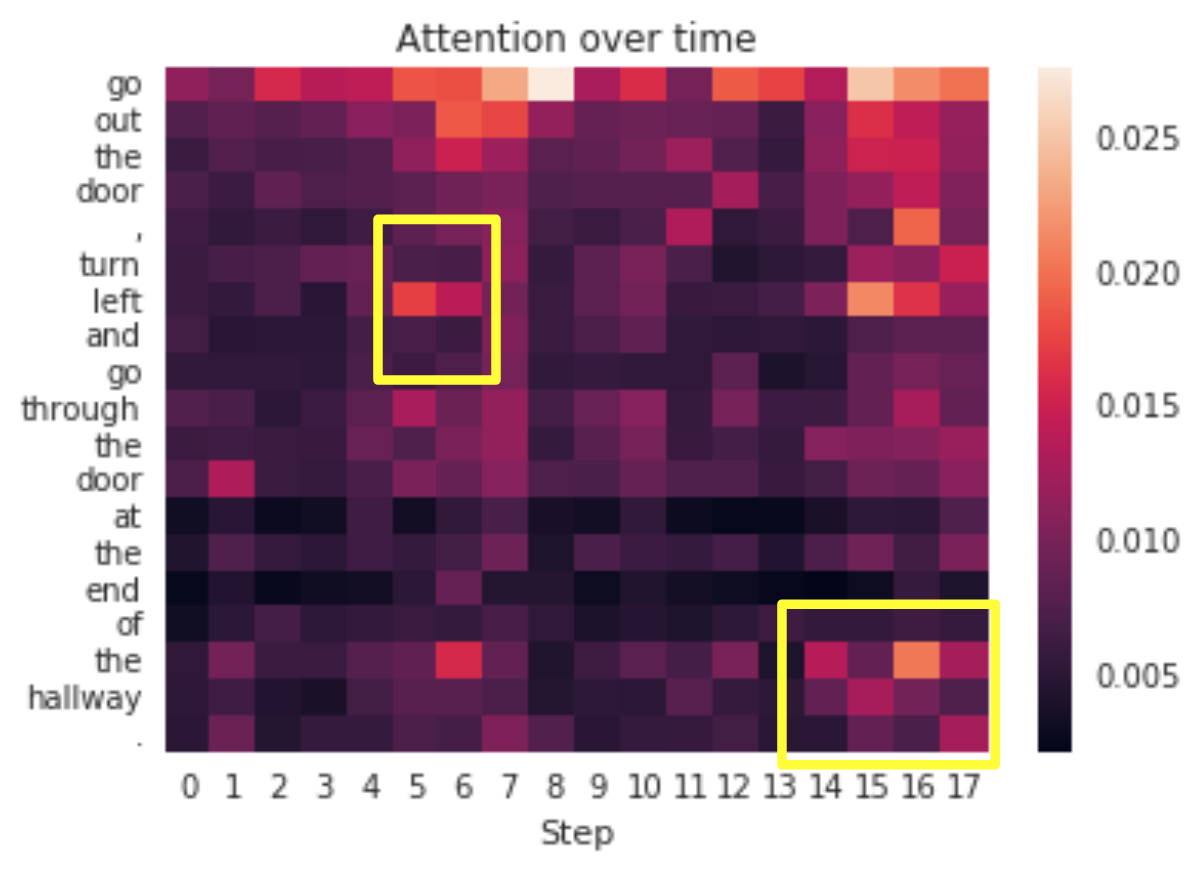}\label{fig:attention_annotated}
		\caption{Language attention heatmap over time steps. Brighter colors indicate more attention. Agent's attention over instruction words shifts based on the agent's location along the path to the goal.}
		\label{fig:attention}
	\end{center}
\end{figure}

\begin{figure*}[]
	\begin{center}
	\includegraphics[width=0.9\textwidth,height=4.3cm,keepaspectratio=false]{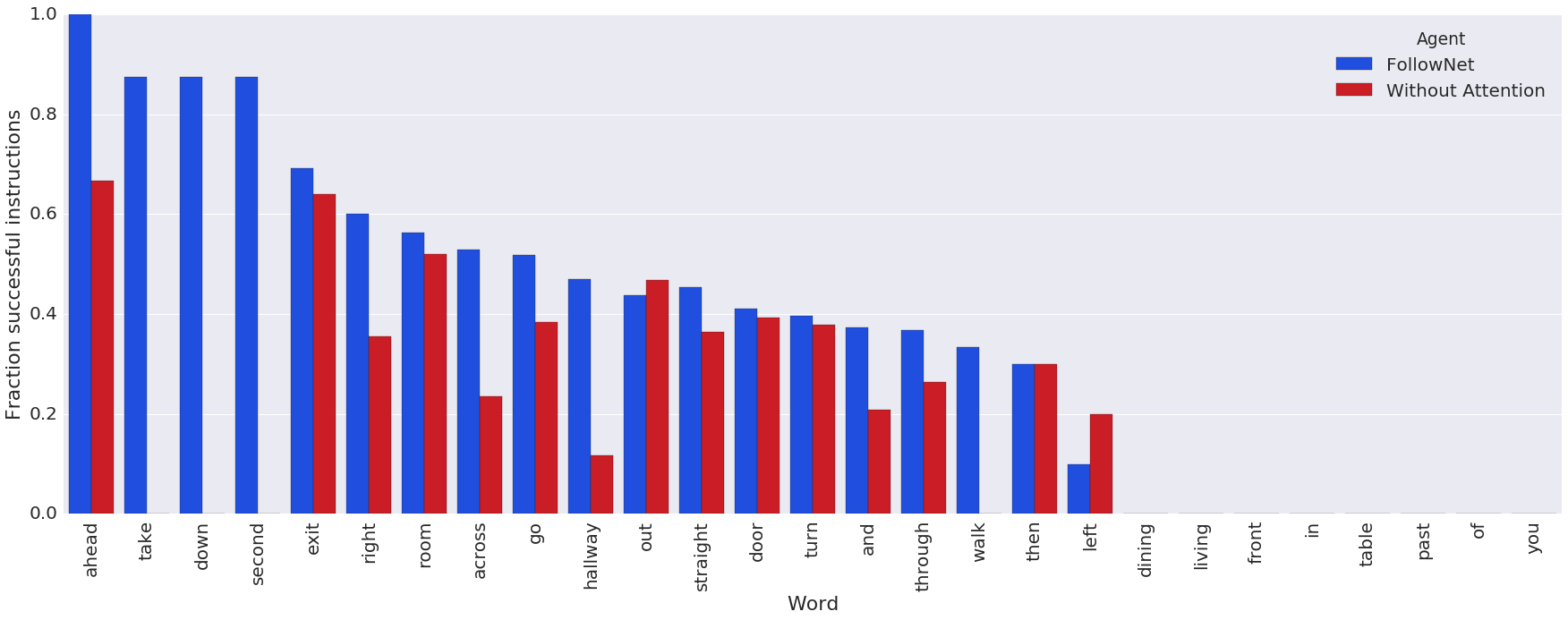}\label{fig:words_att}
		\caption{Fraction of successful tasks per word occurrence on a holdout set.}
		\label{fig:words}
	\end{center}
\end{figure*}

\subsubsection*{Attention Analysis}
Fig. \ref{fig:attention} shows a heatmap of the attention vector over a single episode. Along the y-axis is the tokenized instruction the agent was given. Each word and punctuation mark (commas and periods) is a separate token. Progressing right along the x-axis, at each timestep we see the weight the agent placed on each token, with lighter colors representing higher weights. In steps 5 and 6, the agent increases the attention on the word ``left'' as it takes a left in the environment. Towards the end of the episode, the agent attends to ``the hallway'' as it reaches the end of the task.

Fig. \ref{fig:words} depicts the fraction of successful (fully completed) tasks per word. We see that \net~ is likely to complete tasks with orientation meanings (ahead, take, down, second, left, right, across). The agent without attention generally has the lower success rate across all words, and exhibits slightly different success probabilities across the words. Both agents have difficulty with words that do not appear often in the dataset (past, you), as expected.

\begin{figure}[]
	\begin{center}
			\begin{tabular}{c}
            \subfloat[Steps for successful episodes.]{\includegraphics[width=0.22\textwidth,height=3.cm,keepaspectratio=false]{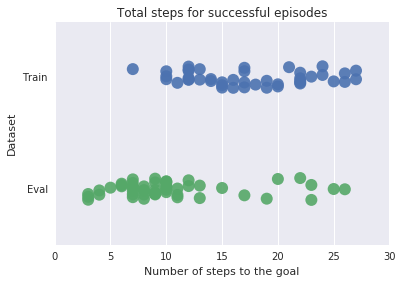}\label{fig:steps_successful}} 

				\subfloat[Fraction of turning actions.]{\includegraphics[width=0.22\textwidth,height=3.cm,keepaspectratio=false]{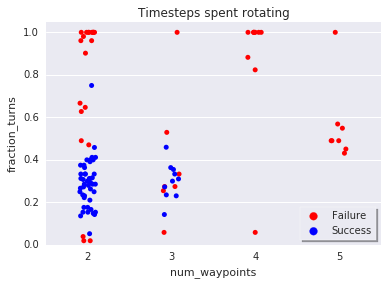}\label{fig:frac_turning}} \\

				\subfloat[Success rate vs. instruction complexity.]{\includegraphics[width=0.3\textwidth,height=3.3cm,keepaspectratio=false]{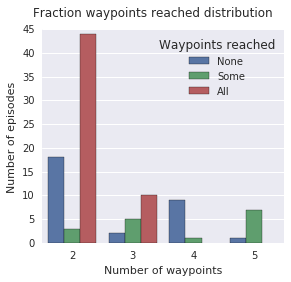}\label{fig:success}}

			\end{tabular}        
		\caption{Motion plan generalization on training instructions. (a) Number of steps per successful episode on the training (blue), and evaluation (green). (b) Fraction of actions that were a left or right turn in successful and unsuccessful episodes, split by number of waypoints in the instruction. (c) Fully (red), partially (green), and not (blue) completed tasks per number of waypoints on the training set.}
		\label{fig:motion}
	\end{center}
\end{figure}

\subsubsection*{Motion plan generalization on complex instructions}
We look now at how well the \net~ agent generalizes the motion plans to new start positions on a complex instruction data set used for the training, up to five-step instructions. Fig. \ref{fig:motion} shows statistics relating to number of steps and number of waypoints in each episode. The training tasks are not trivial, with the number of steps needed to complete the task ranging from 7 to 29 steps, with a mean of 17.4. The evaluation tasks need between 3 and 26 steps, averaging 10.5 (Fig. \ref{fig:steps_successful}).

\net~ agent's overall success at following instructions used in training is \trainsuccess, just 2\% over the the evaluation dataset. This means that the agent generalizes pretty well to the new two-step directions. On the other hand, \trainnosuccess~ of tasks make no progress. It is not surprising that the agent fails more often on the training instructions, because the instructions are more complex. When the agent does not complete the task even partially, it simply spins around without knowing what to do (Fig. \ref{fig:frac_turning}).

Fig. \ref{fig:success} breaks down the episodes by number of waypoints, a proxy for complexity of instructions. The agent is never fully successful at following four- or five-step directions, although in some cases it makes partial progress. Two- and three-step directions are often fully completed. On the evaluation dataset (Fig. \ref{fig:WaypointCompare}), which contains two-step tasks, we notice an interesting bimodal distribution: An agent which reaches the first waypoint is very likely to reach the second. 

\section{Conclusions}
This paper presents the \net~ architecture, which uses an attention mechanism over natural language instructions conditioned on multi-modal sensory observations as an action-value function approximator in DQN. The trained model learns to follow natural language instructions using only visual and depth information. The results show promise that we can simultaneously learn to generalize directional instructions and recognize landmarks. The agent is successful in following novel two-step directions most of the time (at the level of toddler), a 30\% improvement over the baseline. In the future work, we aim to train the agent on a much larger dataset, do more in-depth analysis and empirical evaluation across several domains, and explore generalization across different environments.






\section*{ACKNOWLEDGMENT}
The authors thank James Davidson for the helpful comments and discussions.
\bibliographystyle{abbrv}
\bibliography{literature}


\end{document}